\documentclass{article}

\usepackage{microtype}
\usepackage{graphicx}
\usepackage{subfigure}
\usepackage{amsmath}
\usepackage{amssymb}
\usepackage{booktabs} % for professional tables
\usepackage{xcolor}
\usepackage{multirow}
\usepackage{pifont} % http://ctan.org/pkg/pifont
\usepackage{hhline}
\usepackage{kotex}

\usepackage{hyperref}

\newcommand{\argmax}{\mathop{\mathrm{argmax}}}

\usepackage[accepted]{icml2020}

\icmltitlerunning{A benchmark study on reliable molecular supervised learning via Bayesian learning}

\begin{document}

\twocolumn[
\icmltitle{A benchmark study on reliable molecular supervised learning via Bayesian learning}

% It is OKAY to include author information, even for blind
% submissions: the style file will automatically remove it for you
% unless you've provided the [accepted] option to the icml2020
% package.

% List of affiliations: The first argument should be a (short)
% identifier you will use later to specify author affiliations
% Academic affiliations should list Department, University, City, Region, Country
% Industry affiliations should list Company, City, Region, Country

% You can specify symbols, otherwise they are numbered in order.
% Ideally, you should not use this facility. Affiliations will be numbered
% in order of appearance and this is the preferred way.
\icmlsetsymbol{equal}{*}

\begin{icmlauthorlist}
\icmlauthor{Doyeong Hwang}{AITRICS}
\icmlauthor{Grace Lee}{AITRICS}
\icmlauthor{Hanseok Jo}{AITRICS}
\icmlauthor{Seyoul Yoon}{AITRICS}
\icmlauthor{Seongok Ryu}{AITRICS}
\end{icmlauthorlist}

\icmlaffiliation{AITRICS}{AITRICS, Seoul, Republic of Korea}

\icmlcorrespondingauthor{Seongok Ryu}{seongokryu@aitrics.com}
%\icmlcorrespondingauthor{Eee Pppp}{ep@eden.co.uk}

% You may provide any keywords that you
% find helpful for describing your paper; these are used to populate
% the "keywords" metadata in the PDF but will not be shown in the document
\icmlkeywords{Machine Learning, ICML}

\vskip 0.3in
]

% this must go after the closing bracket ] following \twocolumn[ ...

% This command actually creates the footnote in the first column
% listing the affiliations and the copyright notice.
% The command takes one argument, which is text to display at the start of the footnote.
% The \icmlEqualContribution command is standard text for equal contribution.
% Remove it (just {}) if you do not need this facility.

%\printAffiliationsAndNotice{}  % leave blank if no need to mention equal contribution
\printAffiliationsAndNotice{\icmlEqualContribution} % otherwise use the standard text.

\begin{abstract}
Virtual screening aims to find desirable compounds from chemical library by using computational methods. 
For this purpose with machine learning, model outputs that can be interpreted as predictive probability will be beneficial, in that a high prediction score corresponds to high probability of correctness.
In this work, we present a study on the prediction performance and reliability of graph neural networks trained with the recently proposed Bayesian learning algorithms.
Our work shows that Bayesian learning algorithms allow well-calibrated predictions for various GNN architectures and classification tasks. 
Also, we show the implications of reliable predictions on virtual screening, where Bayesian learning may lead to higher success in finding hit compounds.
\end{abstract}

\section{Introduction}

Predicting molecular properties based on structural input is placed at the heart of computational chemistry. 
Recently, graph neural networks (GNNs), which deal with molecular structure graphs, have been widely used for molecular machine learning.
However, due to the vast size of chemical space and insufficient amount of labeled examples, prediction models often make incorrect predictions on out-of-distribution (OOD) samples. 
A simple way to improve model performance is to acquire more labeled examples, but it demands expensive and time-consuming assay experiments.

Since it is uncertain to determine true label of OOD samples, it is more desirable for models to give predictions with low confidence (predictive probability).
However, neural networks are prone to the problem of over-confident prediction\cite{guo2017calibration}, meaning that their predictive probability (confidence) is usually higher than true correctness.
For example, in virtual screening to find COVID-19 medications, a compound with 0.9 predictive probability will be considered as having 90\% probability of being active, and thus it will be taken into account for experimental validation.
However, over-confident predictions will entail unexpectedly large number of false positive/negative predictions, which discourages the reliability of neural networks.
Thus, evaluating and improving prediction reliability would be essential for successful virtual screening with ML models. 

In that sense, Bayesian learning is an essential choice for virtual screening, which enables to yield reliable neural network models.
Recent advances in Bayesian inference \cite{welling2011bayesian, gal2016dropout, lakshminarayanan2017simple, maddox2019simple} allow practical approximation for computing posterior and Bayesian marginalization, which is long-standing challenges in applying Bayesian learning to neural networks. 
To this end, the works in computer vision tasks with well-established benchmark studies\cite{thulasidasan2019mixup, snoek2019can} has shown that Bayesian learning is beneficial for better generalization to OOD and corrupted samples.
Nevertheless, to the best of our knowledge, there is no benchmark study on Bayesian learning for molecular property prediction tasks concerning prediction reliability. 

In this work, we present a benchmark study on reliable molecular supervised learning with graph neural networks and Bayesian deep learning.
Our work investigate on the effectiveness of the recent Bayesian learning methods on various GNNs and binary classification tasks. 
We observe that most of the methods are helpful, in particular, stochastic weight averaging (SWA) and its variant (SWAG) \cite{izmailov2018averaging, maddox2019simple} show consistently good prediction results.  
Also, we explore models' prediction behaviors by using the histogram of predictive probability in order to help understanding implications of Bayesian inference on virtual screening. 
We have released our codes at \url{https://github.com/AITRICS/mol_reliable_gnn} to assist urgent needs in molecular machine learning, such as discovery of COVID-19 medications.

\section{Backgrounds and Methods}

\subsection{Evaluating prediction reliability} 

Many problems in chemistry applications are given by (binary) classification tasks, e.g. whether the molecule is toxic or not, and whether the molecule is biologically active or not. 
Our goal is to develop classification systems whose output can be interpreted as \textbf{probability} (or \textbf{confidence}) of correct prediction. 
To do so, we evaluate prediction reliability by estimating \textbf{expected calibration error (ECE)} \cite{guo2017calibration}, given by
\begin{equation}
    \text{ECE} = \mathbb{E}_{\text{confidence}}[p(\text{correct}|\text{confidence}) - \text{confidence}].
\end{equation}
Low ECE means that predictive probability value corresponds to true probability of correctness.
We refer to \citeauthor{guo2017calibration} for more precise definition of ECE.

\subsection{Bayesian learning}

A primary goal of Bayesian learning is to infer the posterior distribution $p(w|\mathcal{D})$ of model parameters $w$ given dataset $\mathcal{D}$.
Then, the predictive distribution $p(y^*|x^*, \mathcal{D})$ of output $y^*$ given new input $x^*$ can be computed by Bayesian marginalization:
\begin{equation}
\label{eqn:bayesian_marginalization}
    p(y^*|x^*, \mathcal{D}) = \int p(y^*|x^*, w)p(w|\mathcal{D}) dw
\end{equation}
On the other hand, maximum-a-posteriori (MAP) estimation gives the model parameter $w_{\text{MAP}}$ as the mode of posterior distribution:
\begin{equation}
    w_{\text{MAP}} = \argmax_{w} p(w|\mathcal{D})
\end{equation}
Eq. \ref{eqn:bayesian_marginalization} can mitigate over-confident predictions, which are frequently observed in MAP-estimated models, via marginalizing over all possible model parameters drawn from the posterior. 
Also, predictive uncertainty can be estimated by computing the variance of predictive distribution. 
Note that the predictive uncertainty is given by $\sqrt{\bar{y}(1-\bar{y})}$, where $\bar{y} = \mathbb{E}_{p(y^*|x^*, \mathcal{D})}[y^*]$, for binary classification problems. 

Since exact computation of posterior is mostly intractable for neural networks, 
a variety of approximate Bayesian inference methods have been proposed. 
We consider \textbf{MAP-estimation}, \textbf{Deep Ensemble} \cite{lakshminarayanan2017simple}, \textbf{Monte Carlo dropout (MC-DO)} \cite{gal2016dropout}, \textbf{Stochastic Gradient Langevin Dynamics (SGLD)} \cite{welling2011bayesian}, \textbf{Stochastic Weight Averaging (SWA)} \cite{izmailov2018averaging}, and \textbf{Stochastic Weight Averaging Gaussian (SWAG)} \cite{maddox2019simple} as our baselines to demonstrate effectiveness of Bayesian learning in reliable molecular property predictions.
Further details on our Bayesian learning implementations are provided in Appendix \ref{supp:bayesian_details}.

\subsection{Graph neural networks}

We utilize graph neural networks (GNNs) to handle molecular graph structure inputs.
Various GNN architectures have been proposed, and the recent work by \citeauthor{dwivedi2020benchmarking} has performed ablation studies on node, edge and graph prediction tasks.
We have modified their released code\footnote{\url{https://github.com/graphdeeplearning/benchmarking-gnns}} and implemented Bayesian learning algorithms. 
For our study, we utilize \textbf{Graph Convolutional Network (GCN)} \cite{kipf2016semi}, \textbf{GraphSAGE} \cite{hamilton2017inductive}, \textbf{Graph Isomorphism Network (GIN)} \cite{xu2018powerful}, \textbf{Graph Attention Network} \cite{velivckovic2017graph}, and \textbf{Gated Graph Convolutional Network (GatedGCN)} \cite{bresson2017residual}. 
Further elaboration on implementation is provided in Appendix \ref{supp:gnn_details}.

\section{Experiments}
\label{sec:experiments}

\begin{table*}[h]
\centering
\begin{tabular}{|l|c|c|c|c|c|c|c|c|}
\hline
           & \multicolumn{2}{c|}{BBBP}                & \multicolumn{2}{c|}{BACE}                 & \multicolumn{2}{c|}{HIV}        & \multicolumn{2}{c|}{Tox21}      \\ \hline
           & Single                 & Ensemble        & Single                  & Ensemble        & Single         & Ensemble       & Single         & Ensemble       \\ \hline
None       & 17.9 $\pm$ 4.8         & 15.7 $\pm$ 4.8  & 19.3  $\pm$ 7.0         & 15.4  $\pm$ 5.7 & 1.5  $\pm$ 0.4 & 1.5  $\pm$ 0.3 & 9.6 $\pm$ 1.5  & 8.0 $\pm$ 1.3  \\ \hline
MC-DO & 14.9  $\pm$ 4.8        & 15.5  $\pm$ 3.9 & 13.5  $\pm$ 4.8         & 15.5  $\pm$ 5.3 & 0.9  $\pm$ 0.2 & 1.6  $\pm$ 0.3 & 9.7  $\pm$ 1.5 & 8.6  $\pm$ 1.4 \\ \hline
BBB        & 14.3  $\pm$ 3.7        & 12.9  $\pm$ 2.8 & 12.9  $\pm$ 3.3         & 12.6  $\pm$ 3.6 & 3.0  $\pm$ 0.4 & 2.4  $\pm$ 0.4 & 9.4  $\pm$ 1.4 & 8.4  $\pm$ 1.3 \\ \hline
SGLD      & 14.9  $\pm$ 4.3        & 14.2  $\pm$ 4.7 & 13.1  $\pm$ 3.6         & 12.3  $\pm$ 2.8 & 3.0  $\pm$ 0.4 & 2.5  $\pm$ 0.3 & 9.5  $\pm$ 1.3 & 8.4  $\pm$ 1.3 \\ \hline
SWA        & 7.1  $\pm$ 2.8         & 7.0  $\pm$ 2.8  & 8.6  $\pm$ 1.4          & 8.5  $\pm$ 2.7  & 0.9 $\pm$ 0.2  & 1.2 $\pm$ 0.3  & 3.8  $\pm$ 1.1 & 3.7  $\pm$ 1.0 \\ \hline
SWAG       & 6.9 $\pm$ 2.5 & 7.0 $\pm$ 3.1   & 8.2  $\pm$ 2.0 & 8.8  $\pm$ 2.7  & 1.0  $\pm$ 0.2 & 0.9 $\pm$ 0.3  & 3.7  $\pm$ 1.0 & 3.6  $\pm$ 1.0 \\ \hline
\end{tabular}
\caption{ECE(\%, $\downarrow$) of various Bayesian approaches on BBBP, BACE, HIV, and Tox21 prediction tasks. We report mean and standard deviation of results from eight different experiments with scaffold-splitting.}
\label{tab1:ece}
\end{table*}

\begin{table*}[h]
\centering
\begin{tabular}{|l|c|c|c|c|c|c|c|c|}
\hline
           & \multicolumn{2}{c|}{BBBP}                 & \multicolumn{2}{c|}{BACE}                 & \multicolumn{2}{c|}{HIV}          & \multicolumn{2}{c|}{Tox21}        \\ \hline
           & Single                  & Ensemble        & Single                  & Ensemble        & Single          & Ensemble        & Single          & Ensemble        \\ \hline
None       & 82.7 $\pm$ 6.1          & 85.0 $\pm$ 5.5  & 79.3 $\pm$ 6.4          & 81.7 $\pm$ 5.2  & 75.2 $\pm$ 3.3  & 76.2 $\pm$ 3.0  & 73.5 $\pm$ 4.0  & 75.7 $\pm$ 3.9  \\ \hline
MC-DO & 83.6  $\pm$ 5.7         & 85.1  $\pm$ 5.5 & 80.4  $\pm$ 6.0         & 81.8  $\pm$ 5.0 & 74.6  $\pm$ 2.9 & 76.4  $\pm$ 2.9 & 74.0  $\pm$ 3.9 & 75.5  $\pm$ 4.0 \\ \hline
BBB        & 86.9  $\pm$ 3.7         & 88.2  $\pm$ 3.7 & 81.1  $\pm$ 5.1         & 82.1  $\pm$ 4.7 & 72.9  $\pm$ 3.3 & 74.9  $\pm$ 2.8 & 74.0  $\pm$ 4.2 & 75.2  $\pm$ 4.0 \\ \hline
SGLD      & 85.0  $\pm$ 4.7         & 86.7  $\pm$ 5.3 & 81.2  $\pm$ 5.1         & 81.2  $\pm$ 5.1 & 72.7  $\pm$ 3.4 & 75.0  $\pm$ 2.9 & 73.9  $\pm$ 3.8 & 75.3  $\pm$ 3.9 \\ \hline
SWA        & 91.2  $\pm$ 4.1         & 91.5  $\pm$ 3.4 & 81.4  $\pm$ 3.6         & 81.2  $\pm$ 3.6 & 74.1  $\pm$ 3.0 & 76.2  $\pm$ 2.7 & 78.7  $\pm$ 3.6 & 79.1  $\pm$ 3.5 \\ \hline
SWAG       & 91.2 $\pm$ 4.2 & 91.5  $\pm$ 3.4 & 81.5 $\pm$ 3.6 & 81.2 $\pm$ 3.6  & 73.7 $\pm$ 3.1  & 75.1 $\pm$ 3.0  & 78.8 $\pm$ 3.7  & 79.0 $\pm$ 3.6  \\ \hline
\end{tabular}    
\caption{AUROC(\%, $\uparrow$) of various Bayesian approaches on BBBP, BACE, HIV, and Tox21 prediction tasks. We report mean and standard deviation of results from eight different experiments with scaffold-splitting.}
\label{tab2:auroc}
\end{table*}

In this section, we present the experimental results of using Bayesian algorithms for molecular property prediction tasks with several GNN models.
We elaborate on the details on dataset information -- number of training examples, ratio between positive and negative examples -- in Appendix \ref{supp:datasets}, and training configurations in Appendix \ref{supp:training}. 
Since the molecular datasets in this work are highly sparse and imbalanced, we ran experiments with eight different random seeds and scaffold-splitting \cite{wu2018moleculenet} of each dataset for training, validation, and test.

\subsection{Comparison of Bayesian learning algorithms}

In Table \ref{tab1:ece} and \ref{tab2:auroc}, we show the prediction results for the four different prediction tasks -- BACE, BBBP, HIV, and Tox21 predictions -- where GIN is set as the baseline model architecture and various Bayesian learning methods (Ensemble, MC-DO, BBB, SGLD, SWA, and SWAG) are used. 
We observe that all Bayesian methods are helpful for improving both prediction reliability (lower ECE) and performance (higher AUROC). 
These results indicate that Bayesian learning approaches are beneficial for improving generalization ability in molecular property prediction tasks. 
In particular, SWA and SWAG show superior performance when compared to the other Bayesian approaches for the most cases. 
We provide additional prediction results (Accuracy, Precision, Recall, and F1-score) in Figure \ref{fig:supp:exp1_properties} (see Appendix \ref{supp:exp_results}).

Also, we attempted to check whether the Bayesian learning methods are effective for the other GNN architectures as well as GIN. In Figure \ref{fig:supp:exp2_architecture} (see Appendix \ref{supp:exp_results}), we show the prediction performance and reliability of the five GNN models (i.e. GCN, GIN, GraphSAGE, GAT, and GatedGCN) trained with different Bayesian learning methods on the BACE prediction task.  
We observe that SWA and SWAG consistently show better performance and reliability results than MAP.
On the other hand, other methods show worse results for some GNN models than MAP, for example, Ensemble, MC-DO, and BBB show higher ECE than MAP. 
Specifically, BBB give poor results for GCN and GatedGCN -- showing significantly deteriorated accuracy, recall, and F1-score, despite the fact that we adopted scaling factor to the Kullback-Leibler divergence term in the learning objective of BBB, which can be interpreted cold-posterior, as described in \citeauthor{wenzel2020good}. (see Appendix \ref{supp:bayesian_details} for more details)
We conjecture the reason for such results from the training sensitivity of BBB according to the choice of hyperparameters (e.g. prior length scale).
We leave deeper investigation of BBB on molecular prediction as future work.

\subsection{Using Deep Ensemble additionally improves Bayesian learning}

\citeauthor{wilson2020bayesian} proposed that the ensemble of SWAG (Multi-SWAG), which uses the ensemble of variational posterior in order to model multi-modal posterior, can improve the single SWAG. 
Motivated by the Multi-SWAG, we compare ECE and AUROC results of using single Bayesian models and the Ensemble of Bayesian models on the four prediction tasks, shown in Table \ref{tab1:ece} and \ref{tab2:auroc}.
Using ensemble additionally improve both prediction reliability and performance for all Bayesian approaches in most cases, but its amount is relatively smaller than the improvement gain from using SWA/SWAG. 
Thus, we conclude that using the ensemble of SWA/SWAG would be the best choice to accomplish both high prediction performance and reliability as long as enough computing resource is secured.

\subsection{Prediction behavior of Bayesian models - implications on virtual screening}

\begin{figure*}[h] 
\centering
    \includegraphics[width=0.95\textwidth,trim={0cm 0 0cm 0},clip]{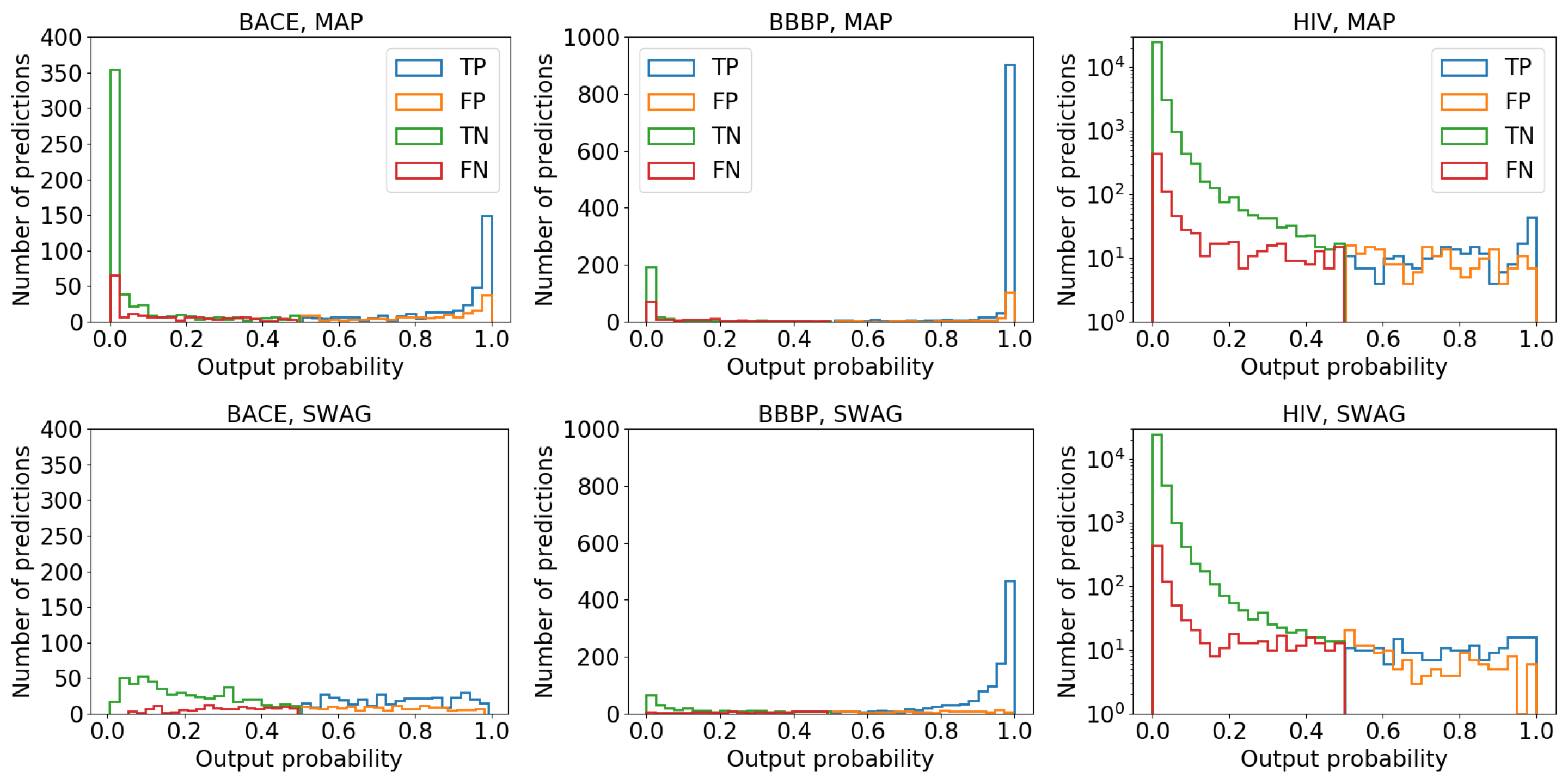}
    \caption{The histograms of \textcolor{blue}{true positive (TP)}, \textcolor{orange}{false positive (FP)}, \textcolor{green}{true negative (TN)}, and \textcolor{red}{false negative (FN)} predictions from the GIN trained with MAP (top) and SWAG (bottom).}
    \label{fig:exp3_behaviour}
\end{figure*}

\begin{figure}[h] 
\centering
    \includegraphics[width=0.48\textwidth,trim={0cm 0 0cm 0},clip]{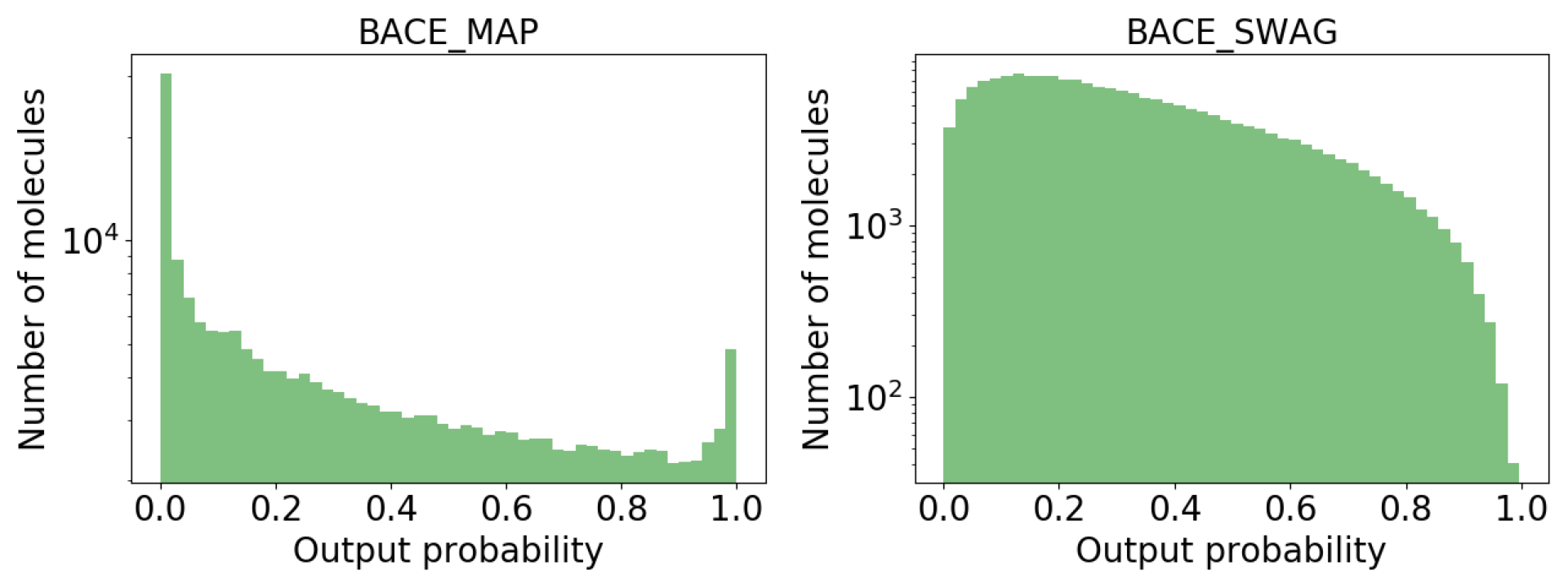}
    \caption{The histograms of predictive probability of BACE-activity from the GIN models trained with MAP (left) and SWAG (right), respectively.}
    \label{fig:exp4_bace_inference}
\end{figure}

Over-confident prediction behavior is frequently observed in neural networks, especially in MAP-estimated models. 
As shown in Figure \ref{fig:exp3_behaviour}, we can confirm that most prediction results from the MAP-estimated models are positioned near zero or one. 
On the other hand, SWAG models effectively mitigate over-confident predictions, which were quantitatively evaluated by using ECE as shown in Table \ref{tab1:ece} -- much smaller number of predictions and higher ratio between true positive/negative and false positive/negative near zero or one.
We show the results obtained with other Bayesian learning methods in Figure \ref{fig:supp:bace_predictions}, \ref{fig:supp:bbbp_predictions}, and \ref{fig:supp:hiv_predictions} (see Appendix \ref{supp:exp_results}).

In addition, we imitated the virtual screening experiment -- using the model trained with the BACE dataset to screen BACE-active compounds from the lead-like subset of ZINC database \cite{irwin2005zinc}.
Though we do not have true label of the compounds in the ZINC dataset, most of them might be out-of-distribution against the samples in the BACE dataset. 
Figure \ref{fig:exp4_bace_inference} shows the histogram of predictive probability for the ZINC compounds inferred by the models trained with BACE dataset. 
We observe that most of prediction results from the MAP model are positioned near zero or one, which might be unnatural for predictions on OOD samples.
On the other hand, the SWAG model shows small number of predictions in which probability is less than $0.05$ or greater than $0.95$ over entire negative or positive predictions, respectively.
In specific, only 238 compounds have predictive probability value greater than $0.95$ out of total $200,000$ samples.
Considering the situation of selecting the experimental candidates using the output probability values, for example selecting compounds with output probability higher than $0.95$, our demonstration implies that Bayesian approaches may be able to give higher success rate than non-Bayesian approach in virtual screening.

\section{Conclusion}

In this work, we have presented the benchmark study on the reliable molecular prediction models developed by using Bayesian learning methods.
Our demonstrations show that the recent Bayesian learning methods are notably beneficial for obtaining well-calibrated prediction results, which would be essential for virtual screening with the final output of neural networks.
We expect our study to be utilized for other applications which can be benefited by the Bayesian principles, such as active learning and continual learning in molecular tasks.

% Acknowledgements should only appear in the accepted version.

\section*{Acknowledgements}
This work was supported by the National Research Foundation of Korea (NRF) grant funded by the project NRF-2019M3E5D4065965.

% In the unusual situation where you want a paper to appear in the
% references without citing it in the main text, use \nocite
% \nocite{langley00}

\bibliography{example_paper}
\bibliographystyle{icml2020}

%%%%%%%%%%%%%%%%%%%%%%%%%%%%%%%%%%%%%%%%%%%%%%%%%%%%%%%%%%%%%%%%%%%%%%%%%%%%%%%
%%%%%%%%%%%%%%%%%%%%%%%%%%%%%%%%%%%%%%%%%%%%%%%%%%%%%%%%%%%%%%%%%%%%%%%%%%%%%%%
% DELETE THIS PART. DO NOT PLACE CONTENT AFTER THE REFERENCES!
%%%%%%%%%%%%%%%%%%%%%%%%%%%%%%%%%%%%%%%%%%%%%%%%%%%%%%%%%%%%%%%%%%%%%%%%%%%%%%%
%%%%%%%%%%%%%%%%%%%%%%%%%%%%%%%%%%%%%%%%%%%%%%%%%%%%%%%%%%%%%%%%%%%%%%%%%%%%%%%
\newpage

\appendix

\section{Backgrounds and implementations of Bayesian learning}
\label{supp:bayesian_details}

This section describes brief backgrounds on Bayesian learning methods.
Also, we provide notes on the implementation of our Bayesian approaches in the following subsections.
We will use the following notations:
\begin{itemize}
    \item $p(\mathcal{D}|w)$ : likelihood function
    \item $p(w)$ : prior on model weights 
    \item $p(w|\mathcal{D})$ : (true) posterior distribution
    \item $q_{\theta}(w)$ : variational distribution, where $\theta$ is the variational parameter. 
\end{itemize}

\subsection{Deep Ensemble \cite{lakshminarayanan2017simple}} 

Deep Ensemble combines the outputs from multiple models, where each of which is trained with different random initialization seed. 
\citeauthor{fort2019deep} analyzed Deep Ensemble as Bayesian approach, showing that each single model in Deep Ensemble corresponds to the different modes of multi-modal posterior distribution and combining the predictive outputs from the different models can be interpreted as Bayesian marginalization.
Furthermore, \citeauthor{wilson2020bayesian} proposed to combine Deep Ensemble with variational methods in order to enhance expressive power on posterior, since variational distribution approximates a single modality of posterior distribution.

We used 10 different random initialization seeds for ensembling in practice.

\subsection{Monte Carlo Dropout (MC-DO) \cite{gal2016dropout}}

MC-DO enables efficient practice of approximate Bayesian learning, whose training and inference procedures do not requires significant modification of standard dropout models.
The predictive probability $\bar{y}^*$ of MC-DO for input $x^*$ is obtained by MC-sampling the final outputs computed by different model weights $w_t$ generated by using stochastic dropout masks:
\begin{equation}
\begin{split}
    \bar{y}^* &= \int p(y^*|x^*, w) q_{\theta}(w) dw \\
    &\approx \frac{1}{T} \sum_{t=1}^{T} p(y^*|x^*, w_t),
\end{split}
\end{equation}
where $T$ is the number of MC-sampling.

For our MC-DO implementation, we used residual dropout in every $l$-th GNN layer:
\begin{equation}
    h_i^{l+1} = h_i^{l} + \text{Dropout}(f^{l}(h_i^{l}, \{h_j^{l}: j \in \mathcal{N}_i \}); p),
\end{equation}
where $f(\cdot)$ is the $l$-th node updating layer (more details will be described in Appendix \ref{supp:gnn_details}) and $p$ is dropout rate. 
We used $p=0.2$ and $T=30$ for our experiments. 

\subsection{Bayes By Backprop (BBB) \cite{blundell2015weight}} 

Variational Bayes aims to minimize the Kullback-Leibler (KL) divergence between the two distributions to model the true but intractable posterior with variational posterior:
\begin{equation}
\begin{split}
    & \text{KL}[ q_{\theta}(w) || p(w|\mathcal{D}) ] \\
    & \geq -\mathbb{E}_{q_{\theta}(w)} \left[ \log p(\mathcal{D}|w) \right] + \text{KL}[ q_{\theta}(w) || p(w) ],
\end{split}
\end{equation}
where the R.H.S is also referred to as evidence lower bound (ELBO).
\citeauthor{blundell2015weight} assumed Gaussian variational posterior and derived the analytic expression of KL-divergence term, leading to obtain variational distribution by using backpropagation. 

We minimized the analytic expression of negative ELBO term in \citeauthor{blundell2015weight}, 
but multiplied a factor of $0.01$ to KL divergence between the variational posterior and the prior, 
which was significantly helpful for training various GNNs with BBB. 
We refer to \citeauthor{blundell2015weight} and \citeauthor{wenzel2020good} for more details on re-weighting of the KL-term.
We used the open-source python library `Blitz - Bayesian Layers in Torch Zoo'\footnote{\url{https://github.com/piEsposito/blitz-bayesian-deep-learning}} for implementing BBB. 
Also, we used single Gaussian prior, while \citeauthor{blundell2015weight} proposed Gaussian mixture prior.

\subsection{Stochastic Gradient Langevin Dynamics (SGLD) \cite{welling2011bayesian}}
\label{supp:sgld}

SGLD can be thought as connecting Monte-Carlo Markov Chain (MCMC) and stochastic gradient descent -- weight transition in MCMC is modeled by the stochastic gradient of posterior.
At time step $t$, the updating rule for weights $w_t$ drawn from posterior is given by
\begin{equation}
    \Delta w_t = \frac{\epsilon_t}{2} \left[ \nabla \log p(w_t) + \frac{N}{n} \sum_{i=1}^{n} \nabla \log p(y_{i,t} | x_{i,t}, w_t) \right] + \eta_t,
\end{equation}
where $\epsilon_t$ is step size (learning rate), $N$ is the number of training examples, $n$ is the size of mini-batches, $(x_{i,t}, y_{i,t})$ is the mini-batch of training examples chosen from the dataset $\mathcal{D}$, and $\eta_t \sim N(0, \epsilon_t)$ is the Gaussian noise. 

For our implementation, we followed the pre-conditioned SGLD\cite{li2016preconditioned}, which is known for more stabilized training with SGLD. 
We did not sample model weights for the first 100 epoches (Burn-in steps), and sampled weight for every two epoch in the sampling steps of 100 epochs after the Burn-in steps. 
%\textcolor{blue}{Also, we set $\epsilon_t = 10^{-3}$. (체크하고 글 수정)} 

\subsection{Stochastic Weight Averaging (SWA) \cite{izmailov2018averaging} and Stochastic Weight Averaging Gaussian (SWA) \cite{maddox2019simple}}
\label{supp:swa_and_swag}

SWA and SWAG sample model weights updated by stochastic gradient descent (SGD) for Bayesian learning, whose theoretical foundation connects the dynamics of weight parameters on loss surface and Bayesian posterior. 
Both algorithms consist of two steps: i) preconditioning step for updating model parameters to the (sub-)optimal point of loss surface , and ii) sampling step for sampling the weight parameters near the (sub)-optimal point generated by SGD optimizer.  

Those two approaches followed the same two steps, but SWA used Polyak-Ruppert weight averaging\cite{polyak1992acceleration} for obtaining the final model weight and SWAG approximates the variational distribution with the mean and covariance of the sampled weights. 

\section{Backgrounds and Implementations of graph neural networks}
\label{supp:gnn_details}

This section describes brief introduction to graph neural networks (GNNs) and our implementations of the GNNs studied in this work. 
We consider molecular graph $G = (V,E)$ whose node features are $x_v = h_v^{0}$ for $v \in V$ and edge features are $e_{ij}$ for $(i,j) \in E$. 

The $l$-th GNN layer updates the $i$-th node features from $h_i^{l} \in \mathbb{R}^{d^{l}}$ to $h_i^{l+1} \in \mathbb{R}^{d^{l+1}}$, where $l \in [0, ..., L-1]$, and its updating rule is given by
\begin{equation}
\label{eqn:node_feature}
\begin{split}
    h_i^{l+1} &= h_i^{l} + \hat{h}_i^{l+1} \\
    &= h_i^{l} + f^l(h_i^{l}, \{(h_j^{l}, e_{ij}) : j \in \mathcal{N}_i\}), 
\end{split}
\end{equation}
where $\mathcal{N}_i$ is the set of nodes adjacent to the $i$-th node.
$f^l(\cdot)$ is the $l$-th node updating layer whose formalism will be described in the following subsections.
In this work, we used same dimension for the all GNN layers' outputs, i.e. $d^{l} = d$ for all $l \in [0, ..., L-1]$.

After applying total $L$ node updating layers, the readout layer aggregates the node features to produce the graph feature $h_G \in \mathbb{R}^{d_G}$:
\begin{equation}
\label{eqn:graph_feature}
    h_{G} = \sum_{v \in V} W_G h_v^{L},
\end{equation}
where $W_G \in \mathbb{R}^{d_G \times d}$ is a weight parameter for linear transformation. 

Finally, the predictive label $\hat{y}_{G}$ is given by
\begin{equation}
    \hat{y}_{G} = W^T h_{G} + b,
\end{equation}
where $W$ and $b$ are the weight and bias parameters of the linear classifier.

\subsection{Graph Convolutional Network (GCN) \cite{kipf2016semi}}

GCN aggregates adjacent nodes' features and multiplies a weight parameter for updating node features:
\begin{equation}
\label{eqn:gcn_original}
\begin{split}
    \hat{h}_{i}^{l+1} &= \textrm{ReLU} (\sum_{j \in \mathcal{N}_i \cup \{i\}} W^{l} h_{j}^{l} ) \\
    &= \textrm{ReLU} (W^{l} \sum_{j \in \mathcal{N}_i \cup \{i\}} h_{j}^{l} ), 
\end{split}
\end{equation}
where $W^{l} \in \mathbb{R}^{d \times d}$ is a weight parameter. 

\subsection{Graph Isomorphism Network (GIN) \cite{xu2018powerful}}

The original node updating formalism of GIN-$\epsilon$ is given by 
\begin{equation}
\label{eqn:gin_original}
\begin{split}
    \hat{h}_{i}^{l+1} &= \text{MLP}( (1+\epsilon^{l}) h_{i}^{l} + \sum_{j \in \mathcal{N}_i} h_{j}^{l} ) \\
    &=  W_2^{l}\text{ReLU}(W_1^{l}( (1+\epsilon^{l}) h_{i}^{l} + \sum_{j \in \mathcal{N}_i} h_{j}^{l} ) ),
\end{split}
\end{equation}
where $\epsilon^{l}$ is the learnable parameter or fixed number, and $W_1^{l}, W_2^{l} \in \mathbb{R}^{d \times d}$ are weight parameters. 
We let $\epsilon^{(l)} = 0$, then the eqn. \ref{eqn:gin_original} is reduced to  
\begin{equation}
\label{eqn:gin_ours}
    \hat{h}_{i}^{l+1} =  W_2^{l}\text{ReLU}(W_1^{l} \sum_{j \in \mathcal{N}_i \cup \{ i \} } h_{j}^{l}  ).
\end{equation}
We can see that the difference between the second line of eqn. \ref{eqn:gcn_original} and eqn. \ref{eqn:gin_ours} is whether using one-layer perceptrons or two-layer perceptrons. 
Note that we did not use batch normalization \cite{ioffe2015batch} for our GIN, in contrast to the implementation in \citeauthor{dwivedi2020benchmarking}. 

\subsection{Graph Sample and Aggregate (GraphSAGE) \cite{hamilton2017inductive}}

Our implementations of GraphSAGE updates the node representations with the following equation
\begin{equation}
   \hat{h}_{i}^{l+1} =  \text{ReLU}(W^{l}\textrm{Concat}[h_{i}^{l}, \sum_{j \in \mathcal{N}_i} h_{j}^{l} ] ), 
\end{equation}
where $W^{l} \in \mathbb{R}^{d \times 2d}$ is a weight parameter. 
We note that the \citeauthor{hamilton2017inductive} proposed mean, sum, max and LSTM aggregation, and we adopted the sum aggregation among them. 
Also, we did not normalize the node features by dividing with their L2-norm, since it can lead to fail graph isomorphism test.

\begin{table*}[h]
    \centering
    \begin{tabular}{|c|c|c|c|c|c|}
    \hline
    \multicolumn{1}{|l|}{}    & BACE          & BBBP         & HIV          & Tox21   \\ \hline
    Task type                 & \multicolumn{4}{c|}{Binary classification}            \\ \hline
    Number of samples         & 1,513         & 2,050        & 41,127       &  7,831  \\ \hline
    Positives:Negatives       & 822:691       & 483:1,567     & 39,684:1,443   &   -     \\ \hline
    Number of tasks & 1             & 1            & 1            &  12     \\ \hline
    \end{tabular}
    \caption{Specifications of the datasets used in this work}
    \label{tab:dataset_spec}
\end{table*}

\subsection{Graph Attention Network (GAT) \cite{velivckovic2017graph}} 

The former GNNs, i.e. GCN, GIN, and GraphSAGE, can be categorized as isotropic node updating methods, in that neighbor nodes' features are aggregated with equal importance. 
On the other hand, GAT adopts multi-head attention mechanism\cite{bahdanau2014neural, vaswani2017attention} for anisotropic node updating, in which neighbor nodes' features are aggregated with learned attention coefficient. 
The updating formalism of GAT is given by
\begin{equation}
       \hat{h}_{i}^{l+1} =  \text{Concat}_{k=1}^{K} \left[ \text{ELU}(\sum_{j \in \mathcal{N}_i} \alpha_{ij}^{k,l} W^{k,l} h_{j}^{l}) \right], 
\end{equation}
where $K$ is the number of attention heads, $W^{k,l} \in \mathbb{R}^{\frac{d}{K} \times d}$, and the attention coefficient $\alpha_{ij}^{k,l}$ is defined as
\begin{equation}
    \alpha_{ij}^{k,l} = \frac{\exp{ \hat{\alpha}_{ij}^{k,l}}}{\sum_{j' \in \mathcal{N}_i} \exp{ \hat{\alpha}_{ij'}^{k,l}}},
\end{equation}
\begin{equation}
    \hat{\alpha}_{ij}^{k,l} = \text{LeakyReLU} ( U^{k,l} \text{Concat} \left[ W^{k,l} h_{i}^{l}, W^{k,l} h_{j}^{l} \right])
\end{equation}
where $U^{k,l} \in \mathbb{R}^{\frac{2d}{K}}$ is a weight parameter. 
Note that we set the number of attention heads as 4.

\subsection{Gated Graph Convolutional Network (GatedGCN) \cite{bresson2017residual}}

GatedGCN is another anisotropic node updating method, which utilizes edge features for updating the edge gates $e_{ij}^{l}$ and multiplies them by the neighbor nodes' feature.
\begin{equation}
    \hat{h}_{i}^{l+1} = \text{ReLU}(U^{l} h_i^{j} + \sum_{j \in \mathcal{N}_i} w_{ij}^{l} \odot W^{l} h_{j}^{l}),
\end{equation}
where $\odot$ denotes the element-wise multiplication, and $U^{l}, W^{l} \in \mathbb{R}^{d \times d}$ are weight parameters.
The gate coefficient $w_{ij}^{l}$ is given by
\begin{equation}
    w_{ij}^{l} = \frac{\sigma(\hat{w}_{ij}^l)}{\sum_{j' \in \mathcal{N}_i} \sigma(\hat{w}_{ij'}^l)}
\end{equation}
\begin{equation}
    \hat{w}_{ij}^{l} = \hat{w}_{ij}^{l-1} + \text{ReLU}(A^{l} h_i^{l-1} + B^{l} h_j^{l-1} + C^{l} \hat{w}_{ij}^{l-1}),
\end{equation}
where $A^l, B^l, C^l \in \mathbb{R}^{d \times d}$ are weight parameters and $\hat{w}_{ij}^{0} = e_{ij}$. 
We again notify that we did not use batch normalization for our implementation of GatedGCN.

\section{Details on molecular datasets}
\label{supp:datasets}

We describe the specifications of the datasets used for training the models in Table \ref{tab:dataset_spec}. We downloaded all the four datasets at the MoleculeNet homepage\footnote{\url{http://moleculenet.ai/datasets-1}}. 
Scaffold-splitting to training, validation and test sets by 80:10:10 ratio is applied for each dataset.
Note that the Tox21 dataset consists of 12 different binary classification tasks, and each task has different number of positive and negative samples. 

\section{Details on model training}
\label{supp:training}

In this section, we describe the hyperparameter settings used for the implementation of GNNs and Bayesian learning methods.

For all GNN models, we used the dimension of node features ($d$ in eq. \ref{eqn:node_feature}) as 128, and the dimension of graph feature ($d_G$ in eq. \ref{eqn:graph_feature}) as 256, and the number of node updating layers $L$ as 4.

Since SGLD, SWA, and SWAG utilize gradient descent update for sampling weights from the posteriors, 
we used different optimizer and learning rate scheduling for different Bayesian learning methods.
For MAP, Ensemble, MC-DO, and BBB, we used Adam optimizer and trained models for 200 epochs with initial learning rate of $0.001$, which is decayed by the factor of $0.1$ at the 80- and 160-th epoch.
For SWA and SWAG, we used SGD optimizer and trained models for 250 epochs with initial learning rate as 0.1. Preconditioning step is set to 150 epochs. the learning rate is constantly dropped to 0.01 from 75 epoch to 150 epoch, in which before cyclic learning rate is applied. Then, as sampling step starts, cyclic learning rate is applied in between 0.01 and 0.001, following the method proposed in \cite{garipov2018loss}. 
Model weights were collected for every 4 epochs during the sampling step. 
For SWAG, scaling factor applied on SWAG posterior covariance is set to 1.0.

For the setting of prior distribution, we adopted explicit Gaussian prior $N(0, 100.0)$ for BBB, and weight decay coefficient of $10^{-4}$ for the others.

Lastly, we sampled 30 model weights and averaged the output from them for Bayesian marginalization in MC-Dropout and SWAG. For BBB, we sampled 5 and 100 model weights for training and evaluating the model.

\section{Additional experimental results}
\label{supp:exp_results}

\begin{figure*}[h] 
\centering
    \includegraphics[width=0.9\textwidth,trim={0cm 0 0cm 0},clip]{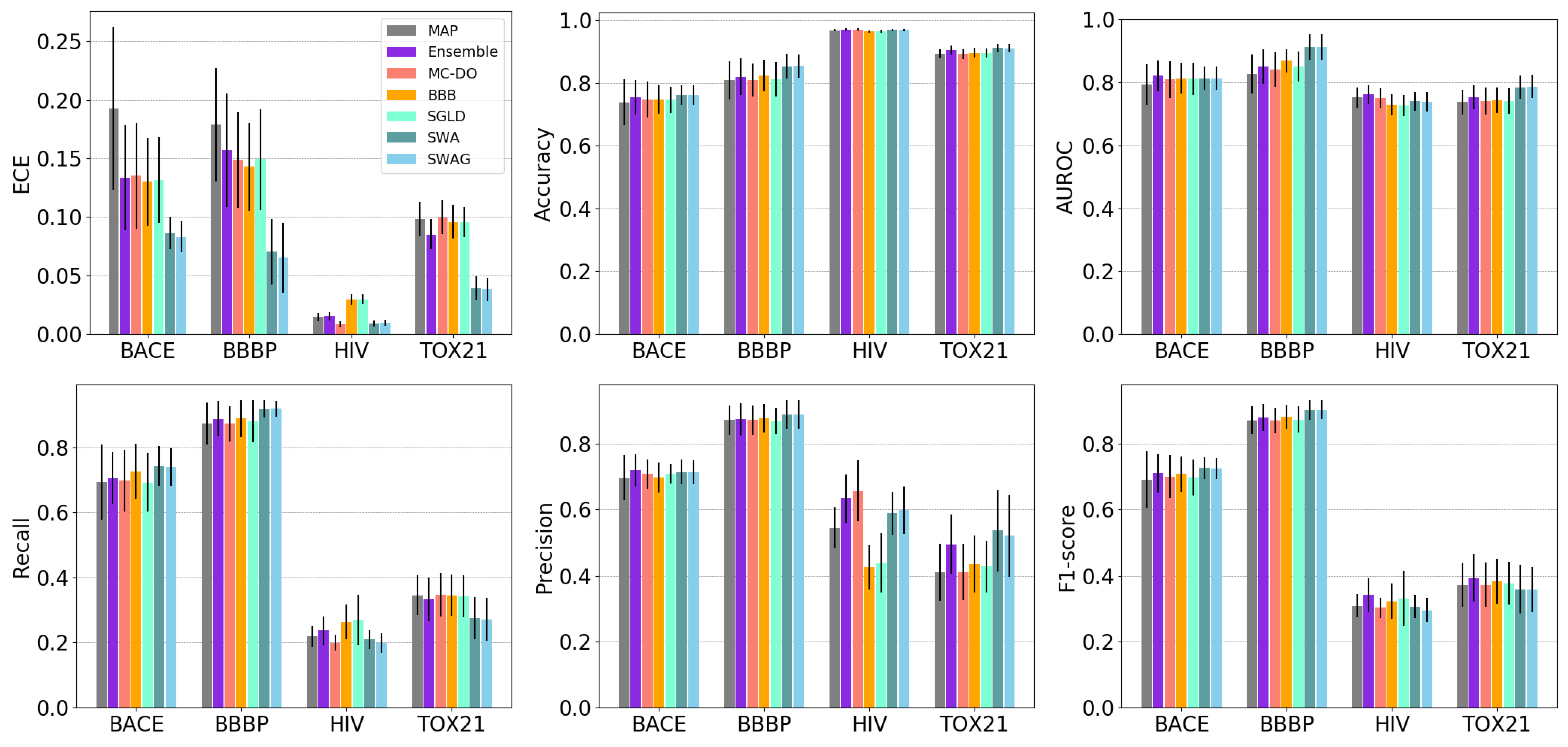}
    \caption{The prediction reliability (ECE; $\downarrow$) and performance (accuracy, AUROC, precision, recall and F1-score; $\uparrow$) of the GIN model for the BACE, BBBP, HIV, and Tox21 prediction tasks. We report mean and standard deviation of results from eight different experiments with scaffold-splitting of the datasets.}
    \label{fig:supp:exp1_properties}
\end{figure*}

\begin{figure*}[h] 
\centering
    \includegraphics[width=0.9\textwidth,trim={0cm 0 0cm 0},clip]{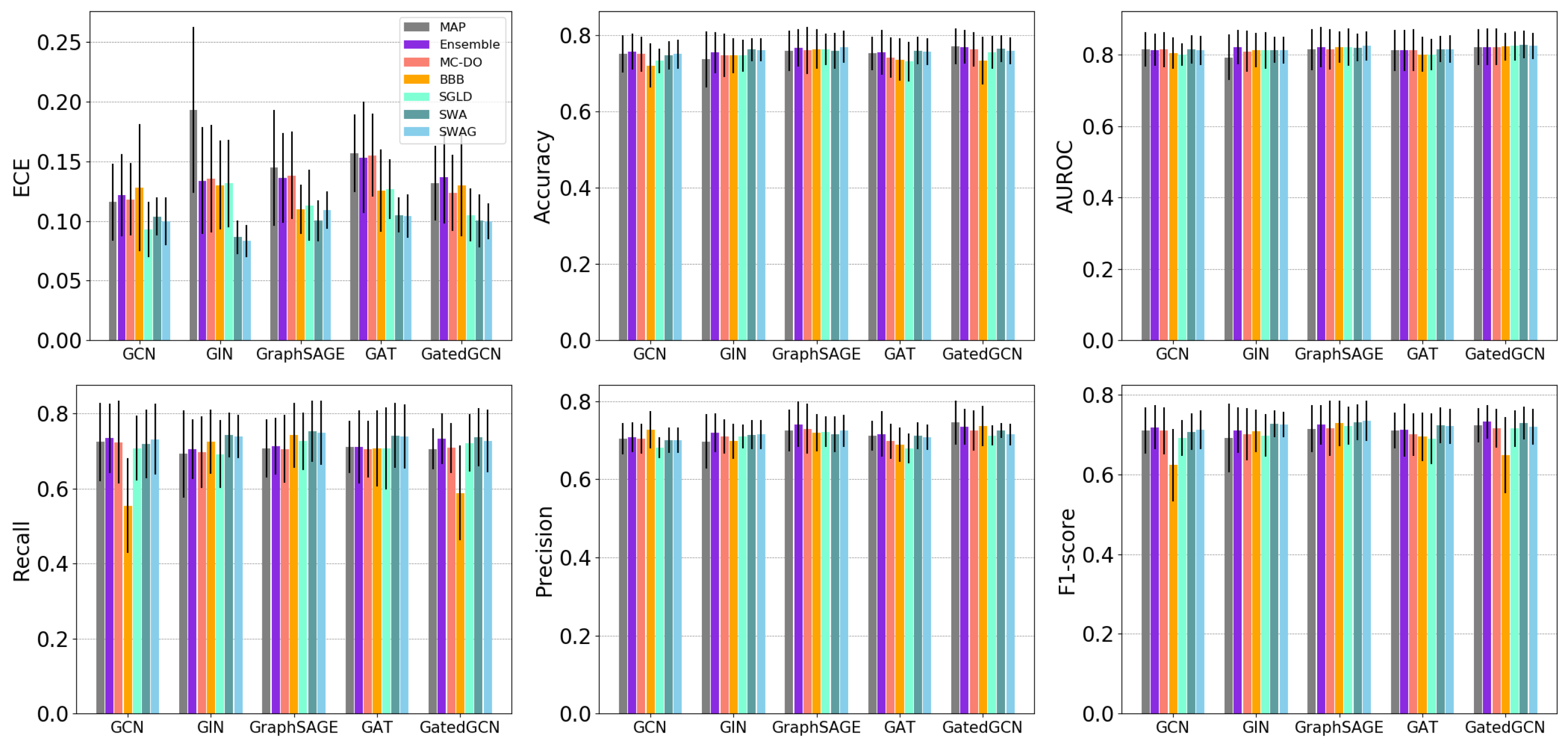}
    \caption{The prediction reliability (ECE; $\downarrow$) and performance (accuracy, AUROC, precision, recall and F1-score; $\uparrow$) of the five different GNN models on the BACE prediction task. We report mean and standard deviation of results from eight different experiments with scaffold-splitting.}
    \label{fig:supp:exp2_architecture}
\end{figure*}

\begin{figure*}[] 
    \includegraphics[width=0.95\textwidth,trim={0cm 0 0cm 0},clip]{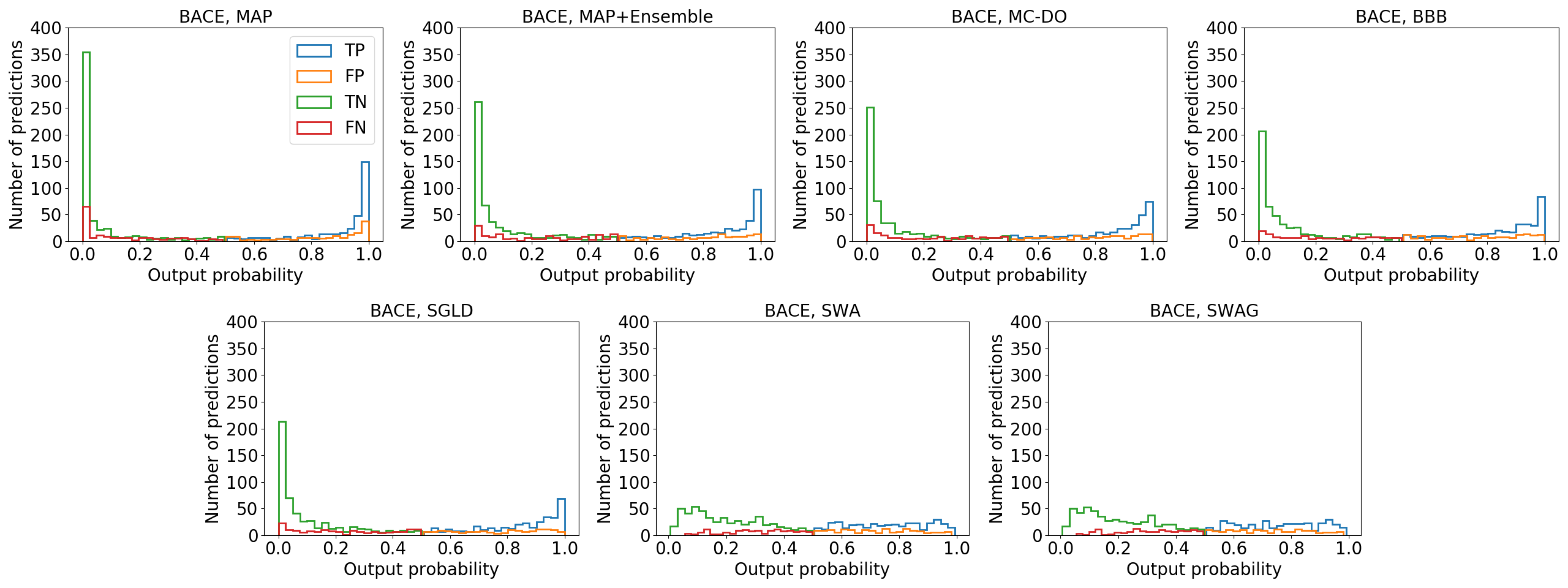}
    \caption{The histogram of \textcolor{blue}{true positive (TP)}, \textcolor{orange}{false positive (FP)}, \textcolor{green}{true negative (TN)}, and \textcolor{red}{false negative (FN)} results for the BACE prediction task.}
    \label{fig:supp:bace_predictions}
\end{figure*}

\begin{figure*}[] 
    \includegraphics[width=0.95\textwidth,trim={0cm 0 0cm 0},clip]{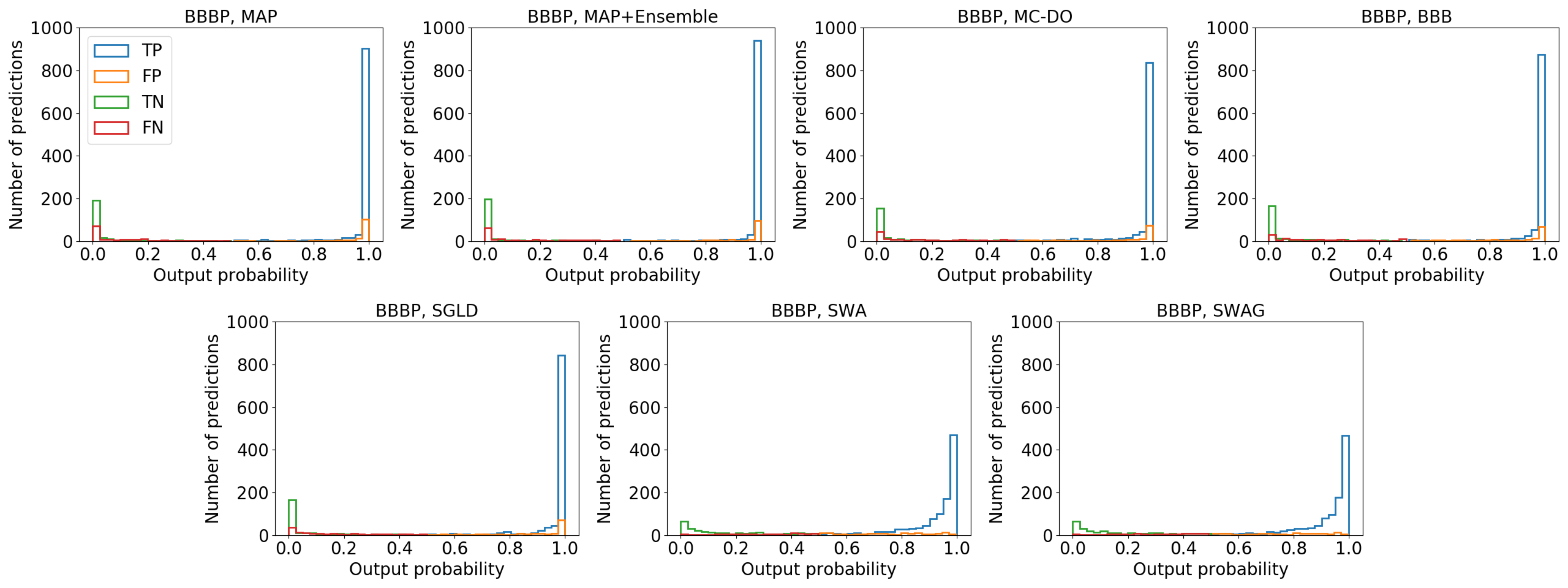}
    \caption{The histogram of \textcolor{blue}{true positive (TP)}, \textcolor{orange}{false positive (FP)}, \textcolor{green}{true negative (TN)}, and \textcolor{red}{false negative (FN)} results for the BBBP prediction task.}
    \label{fig:supp:bbbp_predictions}
\end{figure*}

\begin{figure*}[] 
    \includegraphics[width=0.95\textwidth,trim={0cm 0 0cm 0},clip]{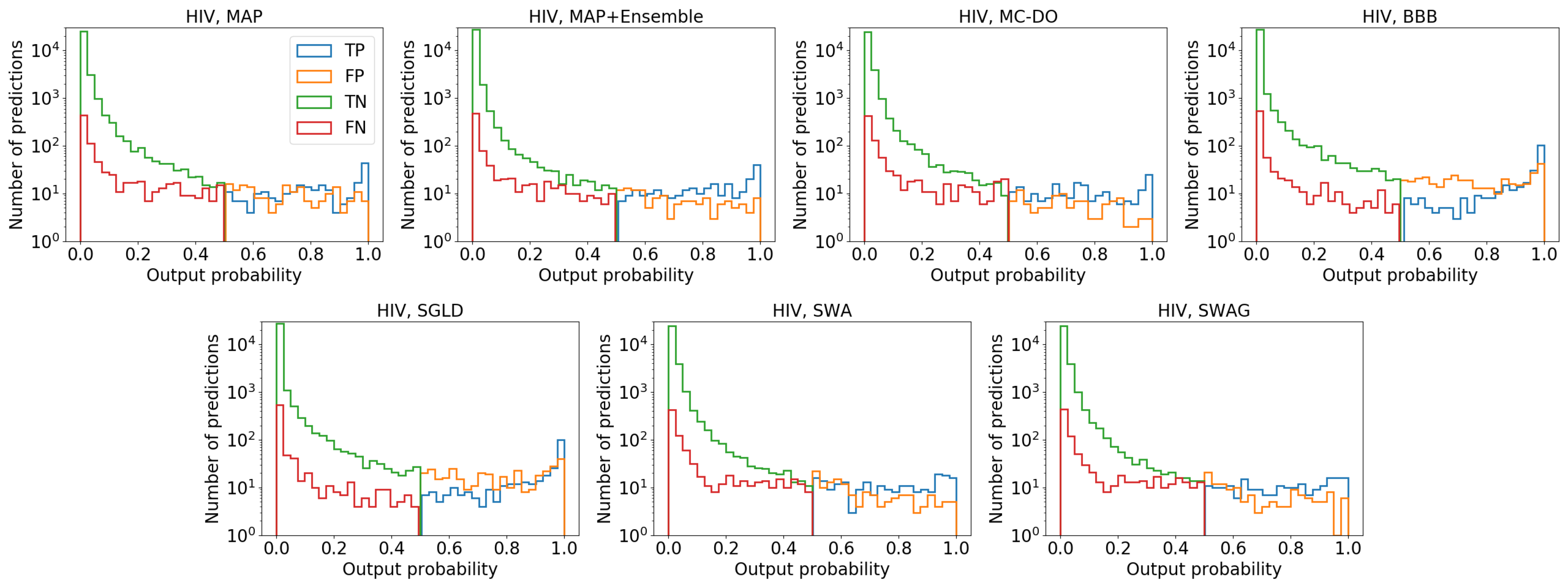}
    \caption{The histogram of \textcolor{blue}{true positive (TP)}, \textcolor{orange}{false positive (FP)}, \textcolor{green}{true negative (TN)}, and \textcolor{red}{false negative (FN)} results for the HIV prediction task.}
    \label{fig:supp:hiv_predictions}
\end{figure*}

In this section, we show the following additional results supporting the main text:
\begin{itemize}
    \item Figure \ref{fig:supp:exp1_properties} shows the prediction reliability and performance of the GIN model for the four molecular property prediction tasks -- BACE, BBBP, HIV, and Tox21 prediction tasks. 
    \item Figure \ref{fig:supp:exp2_architecture} shows the prediction reliability and performance of various GNN models for the BACE prediction tasks.
    \item Figure \ref{fig:supp:bace_predictions}, \ref{fig:supp:bbbp_predictions}, and \ref{fig:supp:hiv_predictions} show the histogram of predictive probability categorized by true positive, false positive, true negative, and false negative predictions for the BACE, BBBP, and HIV prediction tasks. 
\end{itemize}

\iffalse
\section{Do \emph{not} have an appendix here}

\textbf{\emph{Do not put content after the references.}}
%
Put anything that you might normally include after the references in a separate
supplementary file.

We recommend that you build supplementary material in a separate document.
If you must create one PDF and cut it up, please be careful to use a tool that
doesn't alter the margins, and that doesn't aggressively rewrite the PDF file.
pdftk usually works fine. 

\textbf{Please do not use Apple's preview to cut off supplementary material.} In
previous years it has altered margins, and created headaches at the camera-ready
stage. 
%%%%%%%%%%%%%%%%%%%%%%%%%%%%%%%%%%%%%%%%%%%%%%%%%%%%%%%%%%%%%%%%%%%%%%%%%%%%%%%
%%%%%%%%%%%%%%%%%%%%%%%%%%%%%%%%%%%%%%%%%%%%%%%%%%%%%%%%%%%%%%%%%%%%%%%%%%%%%%%
\fi

\end{document}